\pdfoutput=1
\documentclass{article}
\PassOptionsToPackage{hyphens}{url}

\usepackage{microtype}
\usepackage{graphicx}
\usepackage{booktabs}
\usepackage[dvipsnames]{xcolor}
\usepackage{amsmath,amssymb}
\usepackage{mathtools}
\usepackage{enumitem}

\usepackage{hyperref}
\hypersetup{colorlinks=true,linkcolor=teal!60!black,citecolor=teal!60!black,urlcolor=teal!60!black}

\usepackage[accepted]{icml2026}
\makeatletter
\renewcommand{\Notice@String}{Preprint. Preliminary results, shared for discussion; not peer reviewed.}
\makeatother

\newcommand{\R}{\mathbb{R}}
\newcommand{\dstate}{d_{\mathrm{state}}}
\newcommand{\dkey}{d_k}
\newcommand{\dval}{d_v}
\newcommand{\St}{\mathbf{S}_t}
\newcommand{\Stp}{\mathbf{S}_{t-1}}
\newcommand{\armed}{\texttt{armed\_gdn}}
\newcommand{\gdn}{\texttt{gated\_deltanet}}
\newcommand{\dnet}{\texttt{deltanet}}
\newcommand{\dgrank}{\texttt{dg\_rank}}
\newcommand{\dgdiag}{\texttt{dg\_diag}}
\newcommand{\mambaref}{\texttt{mamba2\_ref}}
\newcommand{\mambanoconv}{\texttt{mamba2\_ref\_noconv}}

\icmltitlerunning{Anatomy of Associative Recall in Fixed-State Recurrences}

\begin{document}

\twocolumn[
\icmltitle{Anatomy of Associative Recall in Fixed-State Recurrences:\\
A Matched-State Decomposition, an Interference Wall,\\ and a Curriculum That Breaks It}

\icmlsetsymbol{equal}{*}

\begin{icmlauthorlist}
\icmlauthor{Julian Boesch}{purdue,obit}
\icmlauthor{Andrew Wee}{purdue,obit}
\end{icmlauthorlist}

\icmlaffiliation{purdue}{Purdue University}
\icmlaffiliation{obit}{Obit Research}

\icmlcorrespondingauthor{Julian Boesch}{jboesch@obitmc.com}
\icmlcorrespondingauthor{Andrew Wee}{awee@obitmc.com}

\icmlkeywords{linear attention, state space models, associative recall, MQAR, curriculum learning, diffusion language models, Machine Learning, ICML}

\vskip 0.3in
]

\printAffiliationsAndNotice{}

\makeatletter
\gdef\@icmltitlerunning{Anatomy of Associative Recall in Fixed-State Recurrences}
\makeatother

\begin{abstract}
Fixed-state recurrences---linear attention and state-space models---are widely reported to lag behind attention on associative recall, but whole-architecture comparisons cannot say \emph{which ingredient} is responsible. We decompose masked multi-query recall at a fixed state budget along three single-knob axes: a short causal convolution, the state-transition structure (rank-1 delta rule vs.\ diagonal), and decay. The convolution dominates, adding roughly 0.5 recall accuracy in both families under matched training: comparisons that pit convolution-free cells against a convolution-equipped Mamba are measuring the missing convolution, not the recurrence. The rank-1 transition beats its diagonal ablation by $+0.19$/$+0.32$ at $16$/$32$ pairs, but the margin shrinks to $+0.03$ once both cells carry the convolution, and a state-matched Mamba-2 ties the rank-1 cell before it is ``armed'' with one: no architecture-class claim survives. Cells that solve 32-pair recall degrade gracefully with load yet fall to chance when asked to retrieve just 4 pairs from a haystack of distractors---flat across sequence lengths and transition types. The cause is interference under sparse supervision, not capacity: a \emph{distance curriculum} takes the unchanged architecture from $0.021$ to $1.000$. Training is a \emph{lock-in lottery}---a seed either locks in or does not---and the curriculum is the lever. Lock-in rises from $1/10$ seeds to $7/10$ ($p{=}0.02$); dense supervision adds nothing; at $L{=}256$ a shaped ramp reopens a boundary the uniform curriculum cannot ($4/5$ vs.\ $0/9$); and at $L{=}512$, where the ramp itself collapses ($0/6$), gating it on measured accuracy locks in $6/6$ ($p{=}0.001$). Two side results: bidirectional denoiser cells, which natively read the query before the haystack, show no measurable advantage over causal training at ten seeds, and collision-key retrieval needs two layers. Finally, arming for recall is free on an $S_5$ state-tracking guardrail---the armed cell is significantly \emph{better} at every depth ($p{\le}0.0044$). Together these replace ``recurrent models are bad at recall'' with a measured decomposition and two cheap interventions.
\end{abstract}

\section{Introduction}\label{sec:intro}

Associative recall---binding key--value pairs in context and retrieving a value when its key reappears---is the capability axis on which fixed-state sequence mixers most visibly trail attention. A line of work has made this precise: synthetic multi-query associative recall (MQAR) separates attention from efficient mixers and predicts much of the language-modeling gap between them \citep{arora2023zoology,arora2024based}, and recall remains the standard lens on what a bounded recurrent state can and cannot retain \citep{jelassi2024repeat,hsieh2024ruler}.

But a modern recurrent cell is not one mechanism. A Mamba-2 block, for instance, is a \emph{diagonal selective} state transition \emph{plus} an input-dependent gate \emph{plus} a short causal depthwise convolution \citep{gu2023mamba,dao2024mamba2}; a Gated DeltaNet block is a \emph{rank-1 delta-rule} transition plus a learned decay \citep{yang2024deltanet,yang2024gateddeltanet}; RWKV-7 couples a vector decay with a rank-1 removal term \citep{peng2025rwkv7}. Whole-architecture comparisons therefore confound at least three design axes, and the field's summary judgments (``DeltaNet-style cells recall well,'' ``Mamba recalls better than RNNs,'' ``gating helps'') average over knobs that can be toggled independently. Recent causal-intervention evidence sharpens the concern: Mamba's induction behavior appears to live in its short convolution rather than its state-space scan \citep{arora2025convs,parnichkun2025ess}, though this attribution is itself contested once learning rates are tuned \citep{okpekpe2025recall}. What is missing is a \emph{controlled decomposition}: all knobs, one harness, one fixed state budget.

This paper contributes that decomposition, and then uses it to re-diagnose two failure modes that the decomposition alone does not explain.

\textbf{Contributions.}
\begin{enumerate}[leftmargin=1.4em,itemsep=2pt]
\item \textbf{A matched-state decomposition of recall, and the confound it resolves} (\S\ref{sec:decomp}). At a fixed recurrent-state budget (1{,}024 state elements; ${\sim}29$k parameters) we toggle one knob at a time: short convolution $\times$ transition structure (rank-1 vs.\ diagonal) $\times$ decay. The convolution is the dominant lever and transfers across cell families ($+0.47$ delta-rule, $+0.44$ diagonal at $K{=}32$), corroborating intervention evidence that Mamba's recall is convolution-borne \citep{arora2025convs,parnichkun2025ess}. We also measure the learning-rate sensitivity that \citet{okpekpe2025recall} raise. The rank-1 transition contributes an independent, ablation-grade margin over its own diagonal ablation ($+0.19$/$+0.32$ at $K{=}16$/$32$, 10 seeds, rebinding-controlled), but that margin shrinks to $+0.034$ ($p{=}0.0023$) once both cells carry the convolution, and decay's apparent recall cost dissolves at $20$ seeds ($p{=}0.86$). Two corrections follow. Comparing convolution-free recurrent cells against a convolution-equipped Mamba mismeasures the transition: armed with the same convolution, a gated delta-rule cell reaches $0.99$ at the hardest matched-state setting. The converse also holds. A state- and parameter-matched Mamba-2 ties the \emph{unarmed} rank-1 cell ($0.592$ vs.\ $0.653$, $p{=}0.88$), so no class claim (``rank-1 $>$ diagonal'') survives.
\item \textbf{Capacity vs.\ distance, discriminated} (\S\ref{sec:wall}). Load ($K$) produces graceful degradation; distance across a distractor haystack produces a \emph{wall}: chance already at the shortest tested length---inside the training distribution---identical across delta-rule, selective-diagonal, and RWKV-7 transitions, while a 2-layer attention control solves the task at $1.0$. This signature, completed by the curriculum result below, separates \emph{interference under sparse supervision} from both capacity accounts and decay accounts \citep{echo2026} of recurrent retrieval failure.
\item \textbf{The wall is a training-coverage gap, and training in this regime is a lock-in lottery} (\S\ref{sec:curriculum}). With the architecture unchanged, a table-to-query distance curriculum lifts the walled cell from $0.021$ (chance) to $1.000$---an existence proof by construction. Ten-seed replication reframes the attribution as a \emph{rate} shift and isolates the lever: lock-in $1/10$ under dense supervision alone vs.\ $7/10$ under the distance curriculum ($p{=}0.02$, Fisher two-sided); adding dense supervision to the curriculum changes nothing ($6/10$), and the causal cell locks in at a similar rate ($5/10$). Dense-only training still solves the task outright on one seed. The fix transfers unchanged to $16\times$ state ($d{=}128$). Length is harder: at $L{=}256$ the uniform curriculum fails everywhere ($0/9$ trials) but a shaped ramp locks in $4/5$ seeds ($p{\approx}0.005$, Fisher), and at $L{=}512$ the time-based ramp itself collapses ($0/6$) while a \emph{success-gated} ramp---one that advances the gap only while measured accuracy holds---locks in $6/6$ ($p{=}0.0011$). These are the two curriculum-shape contrasts that clear significance.
\item \textbf{Bidirectional denoisers read twice for free---architecturally; no advantage is measured} (\S\ref{sec:jrt}). The backward stream of a bidirectional masked-denoiser cell sees the query before the haystack, the prefix-encoder property that Just-Read-Twice engineers into recurrent LMs \citep{arora2024jrt}. A collision-key discriminator (decoys reuse the table's keys, defeating any lexical write gate) tests whether the property confers an advantage. At ten seeds it does not: collision training is a bimodal lock-in lottery (bidirectional $3/10$ seeds solve, causal $1/10$; mean difference n.s., $p{=}0.63$), and shaped-curriculum training that stabilizes the lottery ($9/10$ lock-in) does so at exact directional parity. Four things are robust. Both directions can solve the task; any solving circuit needs two layers (a \emph{mark-and-route} circuit, since the one-layer version collapses); the convolution-free cell never locks in even under the shaped curriculum ($0/10$ vs.\ $9/10$ on the identical recipe, $p{\approx}10^{-4}$); and curriculum shape is a powerful direction-agnostic stabilizer ($1/10 \to 9/10$, $p{\approx}6{\times}10^{-4}$).
\item \textbf{A no-cost guardrail} (\S\ref{sec:guardrail}). Arming the cell for recall does not hurt $S_5$ state-tracking learnability; at ten seeds the armed cell is significantly better at every probed depth ($L\in\{2,4,8,16\}$).
\end{enumerate}

All experiments run on a deliberately small, kernel-free, commodity-GPU harness: pure-PyTorch reference cells, a state-matched pure-PyTorch Mamba-2 comparator, and an atomic-free two-pass Triton backward that runs on pre-Volta hardware (Appendix~\ref{app:harness}). That is what makes single-knob matched-state factorials cheap to replicate. The price is scale: every result here is toy-scale ($d\in\{32,128\}$, synthetic tasks), and \S\ref{sec:threats} states the resulting limits explicitly.

\section{Related Work}\label{sec:related}

\textbf{Recall in efficient mixers.} MQAR and its relatives were introduced to explain the attention--SSM gap \citep{arora2023zoology}, and recall-throughput tradeoffs drove a generation of linear-attention designs \citep{arora2024based,katharopoulos2020linear,schlag2021fastweight}. The delta rule \citep{yang2024deltanet} and its gated variant \citep{yang2024gateddeltanet} explicitly target recall via key-conditioned replacement; RWKV-7 generalizes the transition to diagonal-plus-rank-1 \citep{peng2025rwkv7}. Our contribution to this line is not a new cell but a \emph{controlled decomposition} of existing ingredients at matched state, including the negative result that the rank-1 advantage over a diagonal ablation does not extend to a class advantage over Mamba-2.

\textbf{Where Mamba's recall comes from.} Causal interventions locate Mamba's induction in its short convolution \citep{arora2025convs}; effective-state-size analyses likewise find many linear structures collapse on recall without their convolutions \citep{parnichkun2025ess}. \citet{okpekpe2025recall} complicate the attribution, showing tuned learning rates can recover much of Mamba's recall without the convolution---a learnability confound. Our factorial brings a design the intervention studies lack (single-knob toggles at fixed state, in both cell families), though at toy scale. Under matched training we find the convolution's contribution large, additive, and family-transferable, and we measure the learning-rate sensitivity directly (\S\ref{sec:levers}).

\textbf{Long-context retrieval failures.} That SSM-family models fail needle-in-a-haystack retrieval is well documented \citep{hsieh2024ruler,jelassi2024repeat,benkish2024decimamba}; proposed accounts include state capacity and collapse \citep{chen2024stuffedmamba}, finite-horizon state decay \citep{echo2026}, and under-explored state distributions at unseen lengths \citep{lengthgen2025}. Our haystack wall differs from these accounts twice over. Its signature is chance at the \emph{shortest} tested length, flat across lengths, within the training distribution, and transition-independent. Its resolution is a training-signal change alone (\S\ref{sec:curriculum}), which no purely structural account predicts.

\textbf{Training-side fixes.} Birdie improves SSM retrieval with reward-driven objectives and mixed training procedures \citep{blouir2024birdie}; \citet{okpekpe2025recall} and \citet{lengthgen2025} frame recurrent retrieval and length generalization as learnability problems; curricula are classical \citep{bengio2009curriculum}. We add the sharpest version of the claim for retrieval distance: a chance$\to$perfect existence proof with a single-axis curriculum on an unchanged architecture, its lever attribution ($2{\times}2$), and its failure boundary in length.

\textbf{Recurrent denoisers for diffusion LMs.} Masked discrete diffusion \citep{austin2021d3pm,sahoo2024mdlm} makes the denoiser bidirectional by construction; bidirectional Mamba denoisers match Transformer DLMs \citep{singh2025diffumamba}, and RWKV-backbone block diffusion exists \citep{b3drwkv2026}. Just-Read-Twice \citep{arora2024jrt} showed causal recurrent LMs recover most of the recall gap when the query precedes the context. Our contribution is the explicit connection: bidirectional denoisers possess the JRT property \emph{natively}, with no prompt repetition and no bespoke prefix-LM. We add the depth requirement of any circuit exploiting it (two layers: mark-and-route) and its apparent local-binding requirement (the short convolution). Whether the native property confers a measurable \emph{advantage} over matched causal training is open: our ten-seed collision replication finds no significant directional difference (\S\ref{sec:jrt}).

\textbf{State tracking.} Diagonal SSMs and attention are limited on group-composition state tracking in ways rank-1-transition recurrences need not be \citep{merrill2024illusion,grazzi2025negative,liu2023shortcuts,deletang2023chomsky}. We use an $S_5$ word problem only as a \emph{guardrail} (does arming for recall cost state-tracking learnability?) and make no expressivity-class claims here.

\section{Cells, Tasks, and Protocol}\label{sec:setup}

\subsection{The cell zoo}\label{sec:cells}

All cells are implemented in one harness as sequence mixers with identical embedding, readout, and training loops; we call each cell configuration under test an \emph{arm}. Per head, a matrix state $\St \in \R^{\dval\times\dkey}$ is updated by a cell-specific transition; the output is a state readout. We write $k_t, v_t, q_t$ (or $r_t$) for the per-token projections.

\textbf{DeltaNet} (rank-1, no decay) applies the delta rule \citep{schlag2021fastweight,yang2024deltanet}, with input-dependent write strength $\beta_t \in (0,1)$:
\begin{equation}
\St = \Stp\!\left(\mathbf{I} - \beta_t k_t k_t^{\top}\right) + \beta_t\, v_t k_t^{\top},
\quad o_t = \St q_t .
\end{equation}

\textbf{Gated DeltaNet} (\gdn; rank-1 + decay) adds a learned scalar gate $\alpha_t\in(0,1)$ \citep{yang2024gateddeltanet}:
\begin{equation}
\St = \alpha_t\, \Stp\!\left(\mathbf{I} - \beta_t k_t k_t^{\top}\right) + \beta_t\, v_t k_t^{\top}.
\end{equation}

\textbf{RWKV-7 reference} (\dgrank; rank-1 + vector decay) uses the diagonal-plus-rank-1 transition \citep{peng2025rwkv7}:
\begin{align}
\mathbf{A}_t &= \operatorname{diag}(w_t) - (a_t \odot \hat{\kappa}_t)\hat{\kappa}_t^{\top},\\
\St &= \Stp \mathbf{A}_t + v_t k_t^{\top},
\qquad o_t = \St r_t,
\end{align}
with data-dependent vector decay $w_t\in(0,1)^{\dkey}$, removal key $\hat\kappa_t$, and in-context rate $a_t$. Its \textbf{diagonal ablation} (\dgdiag) sets the rank-1 term to zero ($a_t\equiv 0$), leaving $\mathbf{A}_t=\operatorname{diag}(w_t)$: the same cell, same parameter count, one knob.

\textbf{Mamba-2 reference} (\mambaref) is a pure-PyTorch state-matched implementation of the Mamba-2 SSD recurrence \citep{dao2024mamba2}: a per-head scalar selective diagonal scan over an explicit state $\mathbf{H}_t\in\R^{d_{\mathrm{inner}}\times \dstate}$,
\begin{equation}
\mathbf{H}_t = a_t\,\mathbf{H}_{t-1} + x_t\, b_t^{\top}, \qquad y_t = \mathbf{H}_t\, c_t,
\end{equation}
with input-dependent $(a_t,b_t,c_t)$ and the standard short causal depthwise convolution on its input projections. \mambanoconv{} sets the convolution width to 1 (a true toggle). Because this comparator needs neither Triton nor \texttt{mamba\_ssm}, the full factorial runs on commodity GPUs; its cost is that it under-trains relative to the official fused kernel (\S\ref{sec:threats}).

\textbf{The arming knob.} A width-4, strictly causal (left-padded), depthwise convolution on the interaction projections---$(q,k,v)$ for the delta-rule cells, $(v,k,r)$ for the RWKV-7 cells---applied pre-activation exactly as in Mamba-2. The convolution adds ${\approx}0.5$k parameters and \textbf{zero recurrent state}, so state-matching is preserved. \armed{} $\coloneqq$ \gdn{} $+$ this convolution.

\textbf{Attention control.} A standard bidirectional (or causal, where noted) softmax-attention block \citep{vaswani2017attention}, used as the ceiling for the haystack task (\S\ref{sec:wall}) with widened heads (head dimension 16). At the constrained-width factorial setting ($d{=}32$) no valid attention control exists at matched width: the head dimension degenerates, and a two-head variant collapses at one layer for a structural reason (\S\ref{sec:threats}). We therefore do not report attention there.

\textbf{Bidirectional variants.} For \S\ref{sec:jrt}, a cell is made bidirectional in the standard masked-denoiser way \citep{singh2025diffumamba}: a paired backward stream runs the same recurrence on the flipped sequence and the two streams are merged per position. \texttt{armed\_gdn\_fwd} denotes the causal (forward-only) ablation.

\subsection{State matching}\label{sec:statematch}

The factorial is run at \emph{constrained} state ($d{=}32$, one layer), the regime in which recall discriminates between cells. The rank-1 cells carry $\dkey^2 = 32^2 = 1024$ recurrent state elements per head group; the Mamba-2 reference is set to $\dstate{=}16$ so that $d_{\mathrm{inner}}\cdot\dstate = 64\cdot 16 = 1024$ elements---matched state, and matched parameters (${\sim}28.9$k vs.\ ${\sim}29.3$k). At ample state ($d{=}128$) every cell solves every setting and nothing discriminates, which is itself worth reporting: \emph{recall differences among modern cells are a constrained-state phenomenon}.

\subsection{Tasks and floors}\label{sec:tasks}

\textbf{Masked MQAR.} Each sequence is a table of $K$ key--value pairs followed by queried keys whose answers are \emph{single masked tokens}, and the model is trained to fill the masks. Values are drawn fresh at random per sequence from a 26-token range, so no key$\to$value prior exists to memorize. $K\in\{8,16,32\}$. Floors: the value-marginal is $1/26\approx 0.038$; the strongest \emph{no-binding} strategy (emit a random table value) scores ${\approx}0.068$. In the hardened protocol we also report the \emph{table floor} $1/K$ (guess among the $K$ values actually present), the floor the naive marginal comparison misses.

\textbf{Haystack retrieval.} Each sequence is $4$ key--value pairs, then a distractor haystack of task-format tokens, then the query; the answer is a single masked token at the final position. Sequence lengths are $L\in\{64,\dots,512\}$ and chance is $1/54\approx 0.019$. The query-at-end layout makes the task 2-hop; all arms get 2 layers and a proper training budget (the 1-layer version fails for every arm including attention, an under-capacity artifact we exclude).

\textbf{Collision-key variant.} As above, but decoy pairs in the gap \emph{reuse the table's own keys} with fresh random values; ground truth is the \emph{first} (table) binding. This defeats any lexical write gate: distractors are lexically indistinguishable from table keys. Floors: value-marginal $1/54\approx 0.019$; the strongest non-binding \emph{positional} strategy (emit a random table-position token) scores ${\approx}0.26$.

\textbf{$S_5$ guardrail.} A running-product word problem over the symmetric group $S_5$: generators are drawn uniformly from the \emph{full group} (so the answer marginal equals $1/|G| = 1/120 \approx 0.008$ at every depth), each group element is a \emph{single} token, and \emph{all} product slots are masked simultaneously (one parallel fill), so no answer token is completable from a visible prefix. Composition depths $L\in\{2,4,8,16\}$ at $d{=}128$. These three design choices matter: they make the probe structurally immune to the answer-prior and prefix-leakage confounds that can silently invalidate state-tracking probes with multi-token, teacher-forced answers.\footnote{A companion report documents that failure anatomy in a conversion-evaluation setting; this paper's probes are immune by construction, and we state floors with every result.}

\subsection{Statistical protocol}\label{sec:stats}

Exploratory sweeps use 3 seeds. Headline ablations are \emph{hardened}: 10 seeds, two-sided permutation tests on mean gaps, and a \textbf{rebinding control} that evaluates the trained model on sequences whose key--value bindings have been deranged, scoring the \emph{original} value. A genuine retriever's rebinding score must collapse toward the floors (it emits the \emph{rebound} value, not the trained one); a value-salience heuristic does not. For recall and state-tracking probes, these rules operationalize general lessons on prefix leakage in multi-token targets, answer-prior baselines, and control tasks \citep{bachmann2024pitfalls,holtzman2021surface,hewitt2019control}. Where seed outcomes are bimodal (a seed either ``locks in'' or does not), we report \emph{lock-in rates} (fraction of seeds above threshold) with Fisher exact tests rather than means, since mean$\pm$std is meaningless for bimodal outcomes. Every table caption states its seed count, and every task quotes its floors (derivations in Appendix~\ref{app:floors}).

\section{A Matched-State Decomposition of Recall}\label{sec:decomp}

\subsection{The factorial}\label{sec:factorial}

Table~\ref{tab:k32} ranks all seven arms, next to the ingredients each one carries, at constrained state under the hardest load, $K{=}32$.\footnote{No attention control is reported at this setting: at $d{=}32$ with 8 heads the head dimension degenerates to 4, and a two-head variant collapses at one layer because MQAR is an induction-head task that needs two attention layers (\S\ref{sec:threats}). Attention appears as a valid ceiling in the two-layer haystack experiments (\S\ref{sec:wall}), with head dimension 16.} Table~\ref{tab:ksweep} gives the full $K$-sweep (72/72 trials). The next subsection reads both tables one knob at a time.

\begin{table*}[t]
\centering
\caption{\textbf{Constrained state, $K{=}32$ (hardest setting).} Masked recall, mean [min, max] over $n{=}10$ seeds ($n{=}20$ for the two decay arms); these statistics supersede the $n{=}3$ $K{=}32$ column of Table~\ref{tab:ksweep}. State-matched throughout (1{,}024 elements). Floors: value-marginal $\approx 0.038$; no-binding $\approx 0.068$. For context: the official fused Mamba-2 kernel on the same protocol measures $0.867$ at the state-matched $\dstate{=}16$ in our paired session ($0.884$ in the earlier cross-session read) and $1.000$ at $\dstate{=}64$; per-arm tuning lifts the matched-state number to $0.955$ (\S\ref{sec:decomp}).}
\label{tab:k32}
\vskip 0.1in
\small
\begin{tabular}{lccl}
\toprule
arm ($K{=}32$, $d{=}32$) & recall & mean [min, max] & ingredients \\
\midrule
\armed{} & \textbf{0.983} & [0.917, 1.000] & rank-1 $+$ decay $+$ \textbf{conv} \\
\dgrank{} (RWKV-7 rank-1) & 0.653 & [0.507, 0.874] & rank-1 (vector decay), no conv \\
\mambaref{} (state-matched) & 0.592 & [0.196, 0.884] & diagonal $+$ selectivity $+$ conv \\
\dnet{} & 0.561 & [0.094, 0.927] & rank-1, no decay, no conv \\
\gdn{} & 0.518 & [0.296, 0.798] & rank-1 $+$ decay, no conv \\
\dgdiag{} (RWKV-7, rank off) & 0.390 & [0.251, 0.691] & diagonal (vector decay) \\
\mambanoconv{} & 0.151 & [0.079, 0.191] & diagonal $+$ selectivity, no conv \\
\bottomrule
\end{tabular}
\end{table*}

\begin{table}[t]
\centering
\caption{\textbf{Full $K$-sweep}, mean masked recall over $n{=}3$ seeds (72/72 trials); the $K{=}32$ column is superseded by the hardened statistics of Table~\ref{tab:k32} (\dnet{} in particular is bimodal, so its $n{=}3$ mean flatters it). Everything solves $K{=}8$; discrimination grows with load; the no-conv diagonal fails by $K{=}16$; the graceful decline of the unarmed arms from $K{=}8$ to $K{=}32$ is the \emph{capacity} signature---contrast the haystack wall of \S\ref{sec:wall}.}
\label{tab:ksweep}
\vskip 0.1in
\footnotesize
\begin{tabular}{lccc}
\toprule
arm & $K{=}8$ & $K{=}16$ & $K{=}32$ \\
\midrule
\armed{} & 1.000 & 1.000 & \textbf{0.993} \\
\dnet{} & 1.000 & 1.000 & 0.769 \\
\dgrank{} (RWKV-7) & 1.000 & 0.998 & 0.639 \\
\mambaref{} (state-matched) & 1.000 & 0.758 & 0.641 \\
\gdn{} & 1.000 & 0.970 & 0.446 \\
\dgdiag{} & 1.000 & 0.835 & 0.359 \\
\mambanoconv{} & 0.582 & 0.230 & 0.172 \\
\bottomrule
\end{tabular}
\end{table}

\begin{figure*}[t]
\centering
\includegraphics[width=\textwidth]{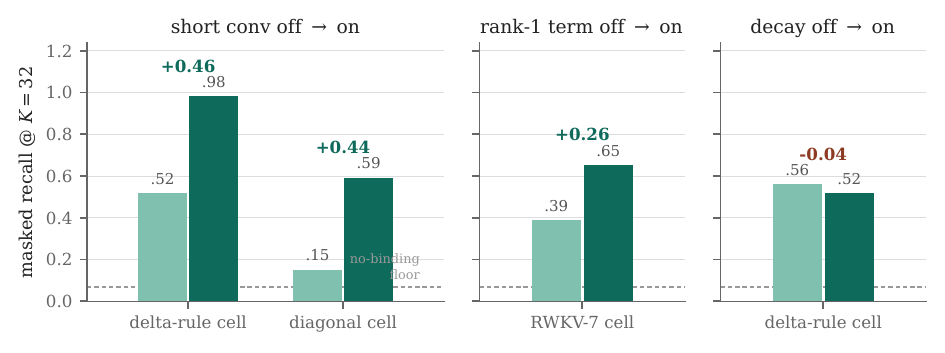}
\caption{\textbf{The three single-knob toggles at $K{=}32$} (constrained state; $n{=}10$ at $K{=}32$ for the five recurrent arms, $n{=}20$ for the decay pair---values from Table~\ref{tab:k32}). Light bars: ingredient off; dark bars: on. The short convolution is the dominant, family-transferable lever; the rank-1 transition adds an ablation-grade within-cell margin (hardened at 10 seeds: $+0.319$ without the convolution, $+0.034$ with it; Table~\ref{tab:hardened}, \S\ref{sec:decomp}); the decay bar's apparent cost dissolves at $n{=}20$ ($p{=}0.86$, \S\ref{sec:decomp}). Dashed line: strongest no-binding floor ($\approx 0.068$).}
\label{fig:decomp}
\end{figure*}

\subsection{Reading the factorial: three separable levers}\label{sec:levers}

Figure~\ref{fig:decomp} shows the three toggles side by side.

\textbf{Axis 1: the convolution ($\approx{+}0.45$, family-transferable).} Hold the recurrence fixed and toggle only the convolution. The delta-rule cell moves $0.518 \to 0.983$ ($+0.47$; \gdn{} $\to$ \armed{}), and the diagonal cell moves $0.151 \to 0.592$ ($+0.44$; \mambanoconv{} $\to$ \mambaref{}). Both legs are true within-cell toggles. The convolution is a \emph{state-free} local primitive, adding no recurrent state, yet it is the largest single lever in the factorial and is of similar magnitude in both families. This is consistent with intervention evidence that Mamba's induction is convolution-borne \citep{arora2025convs,parnichkun2025ess}, and adds a controlled factorial to it.

The size of this lever depends on the tuning regime. We ran the learning-rate check of \citet{okpekpe2025recall} in our own harness (no-conv arms at $\{3\text{e-}4, 1\text{e-}3, 3\text{e-}3\}$, 5 seeds each, rebinding-controlled), and \emph{per-arm tuning rescues the delta family but not our diagonal cell}. Tuned no-conv \gdn{} reaches $0.870$ at lr $3$e-$3$ (vs.\ $0.518$ matched), while \mambanoconv{} peaks at $0.172$ across the sweep, leaving its within-cell convolution gap (${+}0.42$) intact under tuning. Re-tuning the \emph{armed} arms as well, all three saturate at lr $3$e-$3$ (\armed{} $0.999$, armed rank-1 $1.000$, armed diagonal $0.997$; 5 seeds each). The \emph{tuned-vs-tuned} delta-family convolution gap is therefore ${\approx}{+}0.13$ ($0.999$ vs.\ $0.870$), and the with-conv transition margin of ${+}0.034$ is a matched-lr statement, since at tuned lr both armed cells sit at ceiling (margin ${\approx}0.002$). The convolution's contribution is thus \emph{optimization-conditional in the delta family} (large matched, ${\approx}{+}0.13$ tuned) and robust in the diagonal cell; magnitudes should always be quoted with their tuning regime.

\textbf{Axis 2: the transition ($\approx{+}0.3$ without the convolution; $\approx{+}0.03$ with it).} The clean evidence is the within-cell pair \dgrank{} vs.\ \dgdiag{}: one architecture, one parameter count, one switch (the rank-1 term on or off). Without the convolution, at 10 seeds, the pair scores $0.653$ vs.\ $0.390$ ($+0.26$), and the hardened 10-seed rerun (\S\ref{sec:hardened}) confirms it with controls: $+0.191$ at $K{=}16$ ($p{=}0.0002$), $+0.319$ at $K{=}32$ ($p{<}10^{-4}$). The with-conv quadrant is ablation-grade as well (same protocol and harness as Table~\ref{tab:k32}, 10 seeds, rebinding-controlled): the convolution-equipped pair gives $0.992$ (rank-1) vs.\ $0.958$ (diagonal), $+0.034$ ($p{=}0.0023$) at $K{=}32$. At $K{=}16$ all armed arms sit at ceiling ($\ge 0.9998$, 10 seeds), so the margin is a highest-load phenomenon. The transition's contribution is therefore \emph{conditional on the convolution}: large when the cell lacks local mixing, an order of magnitude smaller---though still resolvable---once the convolution is present and both cells operate near ceiling.

\textbf{Axis 3: decay (no measurable cost).} The apples-to-apples comparison within the delta-rule family, at $n{=}20$, is \dnet{} (no decay) $0.561$ vs.\ \gdn{} (decay) $0.518$ at $K{=}32$: $p{=}0.86$, with nothing left of the $-0.32$ the first three seeds suggested. What the seeds do show is bimodality, not a cost. The no-decay cell locks in on $4/20$ seeds (per-seed $0.09$--$0.93$), while the decay cell never exceeds $0.80$ ($0/20$ lock-ins; range $0.30$--$0.80$); decay appears to trade occasional lucky solves for consistency, and neither is a mean-level effect. What is settled is that \armed{} wins with its decay in place, on the strength of the convolution plus the delta rule. The converse claim---that decay buys state-tracking---is likewise not established: a 10-seed lock-in comparison of \gdn{} vs.\ \dnet{} on $S_5$ is directionally positive but non-significant ($p{=}0.65$). One boundary condition comes from the literature: fine-grained (channel-wise) gating can \emph{improve} recall over scalar gating \citep{kimilinear2025}, so any liability measured here is specific to the scalar-decay knob at constrained state.

\subsection{The hardened transition ablation, with controls}\label{sec:hardened}

\begin{table*}[t]
\centering
\caption{\textbf{Hardened transition ablation} ($n{=}10$ seeds; $d{=}32$, $N{=}1$; the three delta-rule arms are convolution-free, while the \texttt{mamba2\_ds16} comparator is a state-matched Mamba-2 that retains its stock short convolution). ``Rebound'' is accuracy on the rebinding control (bindings deranged, original value scored). At $K{=}16$ every arm falls to or below the table floor ($1/K{=}0.062$); at $K{=}32$ rebound sits at $0.042$--$0.073$---slightly above the $1/K{=}0.031$ floor but near the value-marginal ($0.038$) and an order of magnitude below trained accuracy. Models emit the \emph{rebound} value: the recall is genuine key$\to$value binding, not value salience. Permutation tests are two-sided on mean gaps; gaps and drops are computed on full-precision values, not the rounded cells shown.}
\label{tab:hardened}
\vskip 0.1in
\small
\begin{tabular}{llcccc}
\toprule
$K$ & arm & mean [95\% CI] & lock-in $\ge 0.9$ & rebound & drop \\
\midrule
16 & \dgrank{} (rank-1) & 0.998 [0.995, 0.999] & 10/10 & 0.024 & $+0.974$ \\
16 & \dgdiag{} (diagonal ablation) & 0.806 [0.721, 0.892] & 3/10 & 0.038 & $+0.769$ \\
16 & \dnet{} (rank-1) & 1.000 [0.999, 1.000] & 10/10 & 0.024 & $+0.976$ \\
16 & \texttt{mamba2\_ds16} (diagonal, matched) & 0.999 [0.999, 1.000] & 10/10 & 0.024 & $+0.975$ \\
\midrule
32 & \texttt{mamba2\_ds16} (diagonal, matched) & \textbf{0.849} [0.807, 0.890] & 3/10 & 0.042 & $+0.807$ \\
32 & \dgrank{} (rank-1) & 0.702 [0.631, 0.769] & 0/10 & 0.051 & $+0.651$ \\
32 & \dnet{} (rank-1) & 0.623 [0.482, 0.741] & 1/10 & 0.057 & $+0.566$ \\
32 & \dgdiag{} (diagonal ablation) & 0.382 [0.322, 0.450] & 0/10 & 0.073 & $+0.309$ \\
\bottomrule
\end{tabular}
\end{table*}

Table~\ref{tab:hardened} reports the 10-seed, rebinding-controlled rerun of the transition axis. Three findings:

\begin{enumerate}[leftmargin=1.4em,itemsep=2pt]
\item \textbf{The within-cell ablation replicates decisively.} \dgrank{} $>$ \dgdiag{}: $+0.191$ at $K{=}16$ ($p{=}0.0002$) and $+0.319$ at $K{=}32$ ($p{<}10^{-4}$), gaps far larger than seed variance. The rebinding control verifies that what is being measured is binding. The advantage is delta-rule-general, not RWKV-7-specific: \dnet{} $\approx$ \dgrank{} at $K{=}32$ ($p{=}0.36$), both $\gg$ \dgdiag{}.
\item \textbf{No class claim survives.} A state- and parameter-matched \emph{Mamba-2} is a diagonal recurrence, and it \emph{ties} the rank-1 cell at $K{=}16$ ($p{=}0.25$) and \emph{beats} it at $K{=}32$ ($0.849$ vs.\ $0.702$, $p{=}0.0036$). ``Rank-1 class $>$ diagonal class'' is false in this data. The diagonal \emph{ablation} is a crippled member of its class, not a representative; the correct statement is: \emph{within one cell at matched parameters, the rank-1 transition adds recall capacity per state dimension}.
\item \textbf{The two statements together resolve the folklore.} Convolution-free recurrent cells lose to Mamba on recall because of the convolution, not the recurrence (Axis 1). And the rank-1 transition looked alternately strong and weak across studies because its genuine within-cell effect was being compared across architectures against cells that carry the other two knobs.
\end{enumerate}

\textbf{The armed arms at 10 seeds (rebinding-controlled, same harness as Table~\ref{tab:k32}).} \armed{} $0.983$ (per-seed min $0.917$; rebinding $0.037$, at the value-marginal floor), armed RWKV-7 (\dgrank{} $+$ conv) $0.992$ [$0.970, 0.999$], armed diagonal ablation $0.958$ [$0.870, 0.996$]. Within-harness, \armed{} clears \mambaref{} by $+0.392$ ($p{=}10^{-5}$), and the two armed rank-1 families are statistically indistinguishable ($p{=}0.37$): once armed, the delta-rule cell family does not matter. That an armed \emph{diagonal} cell reaches $0.958$ is the factorial's sharpest single number, because it says that at constrained state the convolution is very nearly the whole recall story.

\textbf{The armed cell vs.\ real Mamba, measured in one session.} We ran the official fused Mamba-2 kernel (state-matched $\dstate{=}16$, bf16) through the same probe, session, and Ampere GPU as \armed{}, sweeping each arm's learning rate over a decade ($\{10^{-3}, 3{\times}10^{-3}, 10^{-2}\}$, 5 seeds each). At the shared default ($10^{-3}$) the historical cross-study picture reproduces: official $0.867$ [0.792, 0.911] vs.\ armed $0.967$, so the ${\approx}{+}0.10$ cross-session margin was real at matched lr. Per-arm tuning compresses but does not close it. Both arms peak at $3{\times}10^{-3}$, where the official kernel reaches $0.955$ [0.862, 0.999] and \armed{} $1.000$ on every seed, a tuned-vs-tuned margin of $+0.045$ ($p{=}0.016$, permutation). At $10^{-2}$ \emph{both} arms destabilize (official $0.606$ with two collapsed seeds; armed $0.330$ with four). The reading matches the delta-family result above: most of the official cell's apparent deficit is optimization-conditional, and a smaller armed margin survives tuning. Within our reference harness, \armed{} clears the under-trained \mambaref{} by $+0.39$.

\section{Capacity Curves vs.\ the Interference Wall}\label{sec:wall}

\begin{figure*}[t]
\centering
\includegraphics[width=\textwidth]{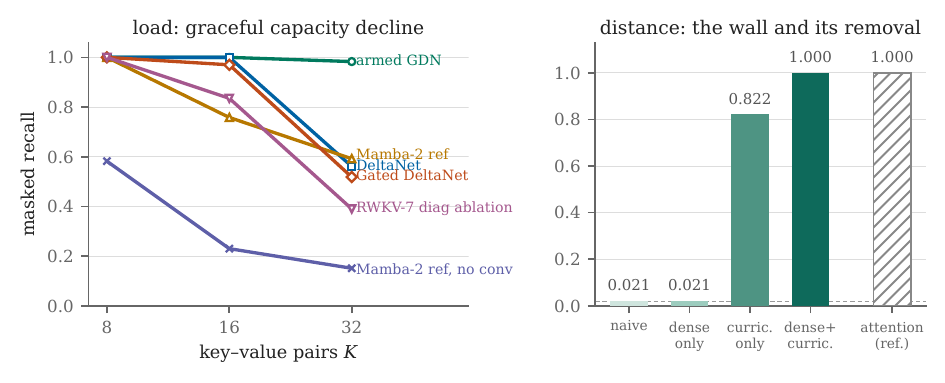}
\caption{\textbf{Load vs.\ distance.} Left: the $K$-sweep (Table~\ref{tab:ksweep}; $n{=}3$ seeds)---a graceful, capacity-style decline; the RWKV-7 rank-1 arm ($0.639$ at $K{=}32$) is omitted for legibility. Right: the haystack wall and its removal for the same bidirectional armed cell at $L{=}64$ (Tables~\ref{tab:wall} and~\ref{tab:curriculum}; bars show seed-0 accuracy, while the latter gives the $n{=}10$ lock-in rates): naive training sits at chance (dashed line, $\approx 0.019$); the distance curriculum unlocks the task. Hatched bar: the attention reference on the fixed layout (\S\ref{sec:curriculum} explains why no attention ceiling is reported under the curriculum itself).}
\label{fig:wall}
\end{figure*}

The $K$-sweep (Table~\ref{tab:ksweep}) shows what a \emph{capacity} limit looks like: everything solves $K{=}8$; the unarmed arms decline gracefully to $K{=}32$. The haystack task shows something categorically different---total, immediate, and length-flat (Table~\ref{tab:wall}); Figure~\ref{fig:wall} juxtaposes the two axes.

\begin{table}[t]
\centering
\caption{\textbf{The haystack wall, flat in length} ($n{=}5$ seeds per cell, means shown; 2 layers, proper budget; chance $\approx 0.019$). The same recurrences that store 32 competing pairs at $\ge 0.99$ (Table~\ref{tab:ksweep}) fail at chance on \emph{4} pairs across a distractor haystack---already at the shortest tested length, for every transition type, with no trend through $4\times$ the length: all 45 recurrent trials lie in $[0.014, 0.023]$, straddling chance. The properly headed attention control is exact at every length.}
\label{tab:wall}
\vskip 0.1in
\small
\begin{tabular}{lccc}
\toprule
arm (2 layers) & $L{=}64$ & $L{=}128$ & $L{=}256$ \\
\midrule
attention (head dim 16) & \textbf{1.000} & \textbf{1.000} & \textbf{1.000} \\
\armed{} & 0.018 & 0.018 & 0.019 \\
\mambaref{} & 0.017 & 0.019 & 0.019 \\
\dgrank{} (RWKV-7) & 0.018 & 0.018 & 0.018 \\
\bottomrule
\end{tabular}
\end{table}

Four properties discriminate the failure mode:

\begin{enumerate}[leftmargin=1.4em,itemsep=2pt]
\item \textbf{It is not capacity.} The same cells store $32$ competing bindings at $\ge 0.99$ (Table~\ref{tab:ksweep}) but fail on $4$ bindings plus distractors. The state is $8\times$ under-subscribed relative to demonstrated capacity.
\item \textbf{It is a wall, not a dip.} The failure is already total at the \emph{shortest} tested length, inside the training distribution, at table-to-query gaps of only tens of tokens, and it stays exactly there through $4\times$ the length: means $0.017$--$0.019$ across $L\in\{64,128,256\}$, with every recurrent trial within $[0.014, 0.023]$ (Table~\ref{tab:wall}). Neither competing account fits that shape. A capacity limit degrades gracefully, as $K$ does; a decay account \citep{echo2026} predicts degradation that grows with distance, not chance at minimal distance and not a floor with no length gradient. There is no distance trend for state decay to explain: the binding is never learned, not progressively lost. (A 1-layer sweep over $L{=}64$--$512$ was likewise flat at chance but is excluded as under-capacity for this 2-hop task.)
\item \textbf{Transition richness does not help.} GDN's delta rule, Mamba's selectivity, and RWKV-7's removal key wall identically---at chance at every tested length ($n{=}5$ each). Whatever is failing is not the state-transition's expressivity.
\item \textbf{Attention is untouched}, as expected, because per-token KV entries cannot be overwritten by later writes: the control is exact at every tested length ($15/15$ trials at $1.000$).
\end{enumerate}

The diagnosis consistent with all four: \emph{interference under sparse supervision}. The model must learn a selective-write policy (store table pairs; ignore distractors), but the training signal---one masked token at the end of the sequence---gives the write gate almost no gradient. Distractor writes overwrite the bindings before the query arrives. This account differs from decay-based explanations of SSM retrieval failure \citep{echo2026}: a decay account predicts distance-dependence and does not predict that a pure training-signal change can fix the failure at fixed architecture. Section~\ref{sec:curriculum} runs exactly that test.

\section{The Wall Is a Training-Coverage Gap}\label{sec:curriculum}

Same task, same architecture, same budget as Table~\ref{tab:wall}; only the training signal changes. Under \emph{dense supervision}, each sequence contains $N$ distinct queried pairs and all of them are supervised. Under the \emph{distance curriculum}, the table$\to$query gap is sampled uniformly in $[0,\text{max}]$ per batch, while evaluation is always at max gap. (Exact flags in Appendix~\ref{app:commands}.)

\begin{table*}[t]
\centering
\caption{\textbf{Curriculum lever attribution, replicated at $n{=}10$ seeds} ($L{=}64$ haystack, 2 layers). Training in this regime converges stochastically (\S\ref{sec:jrt}), so we report lock-in rates (seeds $>0.8$) with per-seed values. The chance$\to$$1.000$ headline is an existence proof by construction and does not rest on seed count; the \emph{attribution} is a rate shift that the curriculum drives: $7/10$ vs.\ $1/10$ ($p{=}0.02$, Fisher two-sided; $6/10$ vs.\ $1/10$: $p{=}0.06$). Dense supervision on top of the curriculum, and the bidirectional--causal contrast, are both n.s. Attention rows: absolute-position $n{=}5$, rotary $n{=}10$.}
\label{tab:curriculum}
\vskip 0.1in
\footnotesize
\setlength{\tabcolsep}{3.5pt}
\begin{tabular}{lcccc}
\toprule
cell ($L{=}64$, 2 layers) & dense & curric. & lock-in & per-seed \\
\midrule
\armed{} (bidir) & 4 & \checkmark & 6/10 & 1.00/0.87/0.88/0.99/0.36/0.45/0.12/0.04/1.00/0.97 \\
\armed{} (bidir) & 1 & \checkmark & \textbf{7/10} & 0.82/0.13/0.06/1.00/0.92/1.00/0.58/1.00/1.00/1.00 \\
\armed{} (bidir) & 4 & --- & 1/10 & 0.02/0.02/0.02/\textbf{1.00}/0.02/0.02/0.02/0.02/0.02/0.02 \\
\texttt{armed\_gdn\_fwd} (causal) & 4 & \checkmark & 5/10 & 0.84/0.75/0.96/1.00/0.06/0.83/0.62/0.24/0.08/0.96 \\
attention (abs.\ pos.) & 4 & \checkmark & 0/5 & all $\approx 0.046$ \\
attention (rotary) & 4 & \checkmark & 6/10 & mean $0.84$; range $0.28$--$1.00$ \\
\bottomrule
\end{tabular}
\end{table*}

\textbf{The existence proof.} The headline row of Table~\ref{tab:curriculum} is an existence proof by construction and needs no seed band to be valid \emph{as an existence claim}: an unchanged 2-layer fixed-state recurrence that sat at chance ($0.021$) under naive training reaches $1.000$ on the same task under a distance curriculum. The wall of \S\ref{sec:wall} is therefore a \emph{training-coverage gap}, not an architectural limit---for this task, at this length. This is the sharpest form of a conclusion that the recent literature reaches by other routes: recurrent retrieval failure is substantially a learnability problem \citep{okpekpe2025recall,lengthgen2025,blouir2024birdie}.

\textbf{Does it survive off the bench?} The curriculum is the one result here that makes a deployment-shaped claim, so it is the one we took to a real converted LM. That work is reported in a companion paper \citep{dreaminggoose}: staged distillation of an attention teacher into a recurrent bidirectional diffusion student, at $1.7$B and $8$B. Here we summarize only its bearing on the claims above. It replicates the phenomenon rather than the number: at conversion scale the curriculum's effect is again a lock-in \emph{rate}. Closing the loop on the curriculum, so that the gap cap advances only while retrieval accuracy holds, raises that rate from one partial lock-in in three to $3/3$ at $0.94$--$0.99$. The recipe carries to $8$B ($2/3$ seeds), with recall transfer rising from $0.000$ to a mean of $0.901$. Two of this section's conclusions therefore hold at four orders of magnitude more parameters: the wall is a training-coverage gap, and training in this regime is a lottery whose rate the curriculum's \emph{shape} controls. One boundary is added there and not visible here: every converted model that solves its trained token band reads chance on held-out token bands, so what transfers is a binding circuit over the symbols it trained on.

\textbf{Lever attribution.} At ten seeds the picture is a family of \emph{lock-in lotteries} whose rates the training signal shifts, and the curriculum is the signal that matters. Dense supervision alone can solve the task (one seed of ten reaches $1.000$) but has the lowest lock-in rate ($1/10$). The curriculum raises it to $7/10$ ($p{=}0.02$, Fisher two-sided; $p{=}0.01$ one-sided), consistent with the mechanism story: training across gaps builds the selective-write policy outward from the solvable adjacent case. Adding dense supervision to the curriculum does not help ($6/10$; vs.\ dense-only, $p{=}0.06$ two-sided), so a denser gradient is not what closes the wall---coverage of the gap is. The causal cell locks in at a similar rate under the same recipe ($5/10$ vs.\ $6/10$, n.s.): directional parity, consistent with the collision replication (\S\ref{sec:jrt}).

\textbf{Two boundaries.} (i) \emph{The attention ceiling under the curriculum requires position generalization---and with it, exists.} The absolute-position toy baseline fails to generalize across table positions when the layout shifts (robust across learning rates and seeds, $7/7$ collapse at ${\approx}0.046$). Replacing its learned absolute table with rotary positions---whose attention scores are provably shift-invariant---restores the ceiling: rotary attention solves the fixed-layout wall ($1.000$) and learns under the shifting curriculum ($6/10$ lock-in lexical, $7/10$ collision-ramp, with most misses graceful rather than collapsed). The absolute-position collapse is thereby confirmed as a position-generalization artifact, not an attention limitation. The shaped-collision recurrent cells' lock-in rate ($9/10$) is statistically indistinguishable from rotary attention's ($7/10$) at these $n$; the regimes are comparable, not ordered.

(ii) \emph{The fix has a length boundary, which shaping re-opens---stochastically.} Width scales cleanly: at $d{=}128$ ($16\times$ state) the $L{=}64$ collision-plus-curriculum recipe reaches $1.000$ unchanged. Length under the \emph{uniform-gap} curriculum does not: at $L{=}256$ (120 colliding decoy pairs vs.\ 24) every uniform cell is at chance, at both widths, at $4\times$ the step budget, at two learning rates. A \emph{shaped} (progressive-ramp) curriculum re-opens it. At five seeds the ramp locks in $4/5$ (per-seed $0.052/0.994/0.996/0.868/1.000$) against $0/9$ pooled uniform trials across widths, budgets, and learning rates, $p{\approx}0.005$ (Fisher, one-sided; the uniform trials pool several configurations). This is the first of two curriculum-\emph{shape} contrasts in this paper that clear significance. So $L{=}256$ is not an architectural interference ceiling; it is the lock-in-lottery phenomenon at a longer horizon, with curriculum shape as a \emph{reliable} rate lever, and the residual lock-out motivating success-gated ramps.

A further $4\times$ in length then exhausts the open-loop ramp outright: at $L{=}512$ every ramp trial sits at chance ($0/6$; bidirectional $0.020/0.020/0.019$, causal $0.023/0.018/0.019$) against $4/5$ at $L{=}256$. Length is thus itself a rate-killer for a \emph{time-based} schedule, which advances on a clock whether or not the model is keeping up. Closing the loop removes the failure. A \emph{success-gated} ramp advances the gap cap only while a running probe-accuracy estimate stays above $0.6$, and never retreats; it locks in $6/6$ seeds at $L{=}512$ (per-seed $0.993$, $0.974$, $0.973$, $0.825$, $0.985$, $0.976$; mean $0.954$) under the same architecture, step budget, and task on which the time-based ramp scored $0/6$ ($p{=}0.0011$, Fisher, one-sided). The $L{=}512$ boundary was therefore never a length ceiling of the two-layer circuit; it was the schedule outrunning the model. A companion paper reports the same closed-loop fix at conversion scale \citep{dreaminggoose}. Both arms are $n{=}6$, and the claim is scoped: gating dissolves a \emph{schedule-borne} boundary, not the token-coverage boundary that paper adds. \textbf{Mean-level results should be quoted as established at $L{=}64$ and, in length, as lock-in rates only.}

\section{Bidirectional Denoisers Read Twice for Free}\label{sec:jrt}

Just-Read-Twice \citep{arora2024jrt} showed that recurrent LMs recover most of the recall gap when the query precedes the context---either by repeating the prompt (JRT-Prompt) or with a bespoke prefix-LM architecture (JRT-RNN). We observe that \emph{bidirectional masked-denoiser cells have this property natively}: the backward stream reads the query before the haystack, by construction, in every bidirectional diffusion-LM denoiser \citep{sahoo2024mdlm,singh2025diffumamba}. No prompt repetition, no bespoke architecture. This section isolates whether that built-in second read is a real mechanism, what circuit implements it, and what it requires.

\textbf{The discriminator.} The lexical haystack cannot settle the question. There the causal cell also lifts under the curriculum ($5/10$ seeds lock in, mean $0.63$; \S\ref{sec:curriculum}, Table~\ref{tab:curriculum}), because distractors are lexically distinguishable from table keys: a causal lexical write gate is learnable in principle, so directionality is a margin there, not a mechanism. The \textbf{collision-key} variant kills that route, since decoys reuse the table's own keys and no lexical gate can separate a distractor write from a table write. The theory going in: an input-conditioned \emph{causal} write gate must overwrite on key reuse (retaining garbage $\to$ chance), whereas the \emph{backward} stream reads the query first and its reverse-scan overwrite retains the correct first-in-forward binding.

\begin{table*}[t]
\centering
\caption{\textbf{Collision-key discriminator, hardened} ($L{=}64$, dense-4 $+$ curriculum, 2 layers; $n{=}10$ seeds, rebinding-controlled---the armed arms' rebound sits at the value-marginal floor, while the convolution-free arm's rebound \emph{equals} its accuracy: it never binds at all). Per-seed accuracies are \emph{bimodal}: most seeds lock out, some solve outright, so we report lock-in rates (seeds $>0.8$) alongside means. The bidirectional--causal mean difference is $+0.082$ ($p{=}0.63$, permutation): \textbf{no directional difference is established at this power}. Floors: value-marginal $\approx 0.019$; strongest non-binding positional strategy $\approx 0.26$.}
\label{tab:collision}
\vskip 0.1in
\footnotesize
\setlength{\tabcolsep}{3.5pt}
\begin{tabular}{lccc}
\toprule
arm ($L{=}64$ collision) & lock-in & mean & per-seed range \\
\midrule
\armed{} (bidirectional) & 3/10 & 0.383 & 0.023--1.000 \\
\texttt{armed\_gdn\_fwd} (causal) & 1/10 & 0.301 & 0.036--0.844 \\
\gdn{} (bidir, \emph{no conv}) & 0/10 & 0.128 & 0.017--0.238 \\
\midrule
\multicolumn{4}{l}{\footnotesize 1-layer ($n{=}5$): both arms chance on every seed (max $0.07$)} \\
\bottomrule
\end{tabular}
\end{table*}

\textbf{Reads (Table~\ref{tab:collision}).}
\begin{enumerate}[leftmargin=1.4em,itemsep=2pt]
\item \textbf{No directional margin at ten seeds.} Both arms are \emph{lock-in lotteries} (bidirectional locks in $3/10$, causal $1/10$; the causal count is threshold-sensitive, since at a $0.6$ criterion both arms lock in $3/10$). Means differ by $+0.082$ ($p{=}0.63$). \emph{``Bidirectional denoisers get Just-Read-Twice for free'' is not established as a differential claim at this power.}
\item \textbf{What does survive: solvability, in both directions.} Some seeds of \emph{both} arms solve collision-key retrieval outright (bidirectional max $1.000$; causal max $0.844$). The causal existence result is itself informative, because it robustly contradicts the single-cell theory that an input-conditioned write gate ``must'' overwrite on key reuse. That strengthens the depth-as-novelty-gate hypothesis: layer-1 state can encode ``this key is already bound,'' making layer-2's gate effectively state-conditioned. We report that as a hypothesis consistent with, but not isolated by, these data.
\item \textbf{The depth requirement is robust.} At one layer \emph{both} arms collapse to chance. The mechanistic reason is exact. The masked answer occupies the \emph{last} position, which is the \emph{first} token of the backward scan, so a single backward pass has absorbed nothing by the time it reaches the answer slot. Query-first information exists only at earlier positions and must be routed to the answer by a second layer's forward stream (\textbf{mark-and-route}). Whatever circuit solves this task, it needs two layers.
\item \textbf{Lock-in is the phenomenon to explain.} The same bimodality appears in the shaped-curriculum length extension (\S\ref{sec:curriculum}: ramp seeds solve $L{=}256$ at $0.994$ or lock out at $0.05$). Curriculum training on interference tasks converges stochastically. The right statistic is therefore a lock-in \emph{rate}; mean accuracy is misleading; and differential claims between arms require far more than ten seeds at these rates ($3/10$ vs.\ $1/10$ is $p{\approx}0.58$ by Fisher exact). The convolution-free arm has never locked in ($0/10$ under the uniform recipe, $0/10$ under the shaped ramp), consistent with the reversed-order-binding account of the convolution's role, but the same power caveat applies.
\end{enumerate}

\textbf{Implication for diffusion LMs.} The architectural observation stands: bidirectional denoisers possess query-first reading natively (the JRT property causal recurrent LMs must engineer via prompt repetition or prefix-LM encoders), and any circuit exploiting it is necessarily $\ge$2 layers (mark-and-route). What these data do \emph{not} show is that the native property confers a measurable advantage over a causal cell trained the same way: at ten seeds the directional difference is indistinguishable from the lock-in lottery. The natural follow-up is whether shaped training that stabilizes lock-in also separates the directions. It does not: under the progressive-ramp curriculum, collision lock-in rises from $3/10$ to $9/10$ (bidirectional) and from $1/10$ to $9/10$ (causal; $p{\approx}6{\times}10^{-4}$ vs.\ its uniform reference), \emph{at exact directional parity} ($p{=}0.76$). The question is closed, negatively: in this task family, query-first reading confers no measurable recall advantage even when training is stabilized. The durable finding is that curriculum \emph{shape} is a powerful, direction-agnostic lock-in stabilizer---the second shape contrast in this paper to clear significance.

\section{Guardrail: Arming Is Free on State Tracking}\label{sec:guardrail}

Does adding a local (finite-window) convolution cost the recurrence anything on the axis where recurrences are supposed to shine---state tracking? We compare \armed{} vs.\ \gdn{} on the $S_5$ running-product guardrail (\S\ref{sec:tasks}).

\begin{table*}[t]
\centering
\caption{\textbf{$S_5$ guardrail} ($d{=}128$; chance $\approx 0.008$; internally comparable only). All four rows are ten-seed (mean [min, max], permutation $p$). The armed cell is significantly \emph{better} at every probed depth: arming for recall does not trade away state tracking. The result is cell-family-general: an armed RWKV-7 cell also beats its unarmed counterpart at both depths ($+0.053$, $p{=}0.004$ at $L{=}8$; $+0.005$, $p{<}10^{-4}$ at $L{=}16$) and is statistically indistinguishable from armed \armed{} ($p{=}0.60$/$0.10$).}
\label{tab:s5}
\vskip 0.1in
\footnotesize
\setlength{\tabcolsep}{3.5pt}
\begin{tabular}{lccc}
\toprule
$S_5$ depth & \armed{} & \gdn{} & $p$ \\
\midrule
$L{=}2$ ($n{=}10$) & \textbf{1.000} [1.000, 1.000] & 0.980 [0.945, 1.000] & $0.0002$ \\
$L{=}4$ ($n{=}10$) & \textbf{0.524} [0.492, 0.637] & 0.389 [0.274, 0.514] & $0.0014$ \\
$L{=}8$ ($n{=}10$) & \textbf{0.214} [0.143, 0.261] & 0.152 [0.132, 0.240] & $0.0044$ \\
$L{=}16$ ($n{=}10$) & \textbf{0.087} [0.072, 0.128] & 0.071 [0.070, 0.073] & $<10^{-4}$ \\
\bottomrule
\end{tabular}
\end{table*}

We frame this strictly as \emph{learnability at fixed depth}, not as a circuit-complexity statement: accuracy falls steeply with composition depth for both arms, and nothing here bears on asymptotic separations \citep{merrill2024illusion,grazzi2025negative}. The design question was narrow---does the recall fix tax the state-tracking axis? The answer is no: the armed cell wins significantly at all four depths.

\section{Limitations and Threats to Validity}\label{sec:threats}

\begin{enumerate}[leftmargin=1.4em,itemsep=2pt]
\item \textbf{The Mamba-2 reference under-trains the official kernel.} \mambaref{} reaches $0.592$ at 10 seeds ($0.641$ at 3) where the official fused Mamba-2 reached $0.884$ at the same matched state. The \emph{within-harness} factorial is internally valid (all cells, one harness, one budget), but absolute margins against a real Mamba are softer. The with-conv transition gap is measured ablation-grade within one cell ($+0.034$, \S\ref{sec:decomp}), so it does not depend on this comparator. The paired measurement is now done in one session with a per-arm learning-rate sweep (\S\ref{sec:decomp}): at the shared default the cross-session numbers reproduce ($0.967$ vs.\ $0.867$), and at each arm's best lr the margin is $1.000$ vs.\ $0.955$ ($p{=}0.016$). Most of the cross-session gap was the official arm's learning rate, and a significant smaller margin survives tuning.
\item \textbf{Toy scale, single task family.} $d\in\{32,128\}$, one MQAR configuration per setting, synthetic tasks. The decomposition's magnitudes ($+0.5$ convolution; $+0.3$ transition without the convolution, $+0.03$ with it; decay unresolved) are specific to constrained state; at ample state everything saturates and the knobs stop mattering. Claims should be read as mechanism isolation, not deployment guidance.
\item \textbf{Learning-rate sensitivity.} \citet{okpekpe2025recall} show per-arm learning-rate tuning can rescue no-conv Mamba recall; our sweep (\S\ref{sec:levers}) confirms this \emph{for the delta family} (tuned no-conv \gdn{} $0.870$; re-tuning the armed arms too narrows the delta-family gap to ${\approx}{+}0.13$) and not \emph{for our diagonal cell} ($0.172$ at its best lr). The sweeps span one decade at 5 seeds; the official kernel's own tuned optimum is $0.955$ (\S\ref{sec:decomp}).
\item \textbf{The attention control is valid only where reported.} Running attention (absolute and rotary) at a healthy head dimension (2 heads, $d{=}32$) through the factorial's matched one-layer config yields collapse at every $K$ (${\approx}0.11$--$0.17$) with the rebinding control \emph{equal} to eval accuracy---the fingerprint of never reading bindings. This is structural. MQAR is an induction-head task and induction circuits require two attention layers \citep{olsson2022induction}, while the recurrent cells solve it at depth one by writing bindings into state. A depth-matched attention row therefore cannot exist at $L{=}1$, and the valid attention ceilings live in the two-layer regimes, where we report them (fixed wall $1.000$; lexical curriculum $6/10$; collision-ramp $7/10$; $L{=}256$ ramp $4/5$---parity with the armed cell's $4/5$). This paper makes no attention-superiority claims beyond those settings.
\item \textbf{Cross-family magnitude comparisons are indicative.} \mambaref{}'s decay is scalar-per-head, the RWKV-7 reference uses vector decay, and \gdn{} uses scalar decay. Within-family toggles are exact; cross-family magnitude comparisons (e.g., convolution effect size in delta vs.\ diagonal families) carry a parameterization caveat.
\item \textbf{Interpretation boundaries we respect.} No rank-1-vs-diagonal \emph{class} claim (\S\ref{sec:hardened} kills it); no claim that decay helps state-tracking (underpowered); no length extrapolation of the curriculum fix beyond $L{=}64$; no circuit-complexity claims from the $S_5$ guardrail; no bidirectional-vs-causal advantage claim under collision (n.s.\ at 10 seeds); the depth-novelty-gating account of the causal arm's solvability is a hypothesis.
\end{enumerate}

\section{Discussion}\label{sec:discussion}

\textbf{What ``recurrent models are bad at recall'' decomposes into.} At matched state, it decomposes into five things: a missing convolution (large, fixable for free in state terms), a transition effect (real, within-cell), a decay tax (real, scalar-gate-specific), an interference failure under sparse supervision (fixable by curriculum, within a length boundary), and a directionality property (bidirectional denoisers get query-first reading natively). None of these is ``the recurrence cannot bind.''

\textbf{When is a hybrid actually needed?} The results re-scope the standard ``add attention layers for retrieval'' move. Two of the classic triggers dissolve under cheap interventions: the unarmed-cell recall deficit (add the convolution) and the within-capacity haystack wall (train with a distance curriculum). Three things remain as genuine hybrid territory: loads beyond state capacity (the graceful $K$-limit), lengths beyond the curriculum boundary ($L{\ge}256$ here, pending shaped curricula), and query-conditioned regimes a fixed-state writer cannot anticipate. A hybrid decision made \emph{after} arming and curriculum is a different, and smaller, decision.

\textbf{For recurrent diffusion LMs.} The JRT-for-free result gives bidirectional recurrent denoisers \citep{singh2025diffumamba,b3drwkv2026} a principled recall story that causal recurrent LMs lack: the objective itself buys the second read. The requirements are concrete---two layers minimum and a short convolution in the backward stream---which makes them design guidance for this model class rather than folklore. The companion paper takes the curriculum half of this story to converted diffusion LMs at $1.7$B and $8$B \citep{dreaminggoose}; whether the query-first mechanism itself carries to scale is untested.

\textbf{A conjecture, and a design question.} Within this study, one apparent architectural wall dissolved entirely into a training-distribution problem, and convergent evidence points the same way for recurrent recall and length generalization \citep{okpekpe2025recall,lengthgen2025,blouir2024birdie}. We state the strong form as a conjecture worth falsifying: \emph{within the capacity regime of a fixed-state recurrence, apparent retrieval walls are training-coverage gaps}. The length boundary of (ii) above is its first direct test---the cell that sits at $0/6$ under a time-based ramp at $L{=}512$ locks in $6/6$ once the ramp is gated on measured competence, so that boundary was the schedule, not the circuit. What remains open against the alternative (a genuine interference ceiling of the two-layer circuit) is whether some length defeats the gated schedule too. Separately, decay costs $-0.32$ on recall here while showing \emph{no measured} state-tracking benefit at our power ($p{=}0.65$); context-modulated forgetting \citep{kimilinear2025} becomes the interesting design axis only if a benefit emerges under better-powered tests.

\section*{Reproducibility Statement}

All cells, tasks, and controls run in one open harness (pure-PyTorch reference cells; a state-matched, kernel-free Mamba-2 comparator; fp32 paths) on commodity GPUs, including pre-Volta hardware via an atomic-free two-pass Triton backward (Appendix~\ref{app:harness}). Appendix~\ref{app:commands} lists the exact commands for every table. Code and result JSONs are publicly available at \url{https://github.com/JIBSIL/dualgoose}.

\section*{Acknowledgments}
This work used computing resources provided by the Rosen Center for Advanced Computing (RCAC) at Purdue University~\citep{McCartney2014}.

\section*{Impact Statement}

This paper presents controlled, small-scale measurements aimed at advancing the scientific understanding of sequence-model architectures: which ingredients of efficient recurrent cells support in-context recall, and when training rather than architecture is the binding constraint. The main downstream impact, if the findings transfer to scale, is more capable linear-time language models at lower inference cost and energy. We see no societal consequences specific to this work beyond those generic to improving machine learning methods.

\bibliography{references}
\bibliographystyle{icml2026}

\newpage
\appendix
\onecolumn

\section{The Commodity-Hardware Harness}\label{app:harness}

Three engineering choices make the factorial cheap to replicate:

\paragraph{Kernel-free cells.} Every arm has a pure-PyTorch reference implementation; the Mamba-2 comparator (\texttt{TinyMamba2RefSSM}) implements the SSD recurrence with an explicit $\dstate$ and requires neither Triton nor \texttt{mamba\_ssm}, so the full factorial executes on any CUDA GPU (and, slowly, CPU). The \texttt{kernel\_size=1} setting neutralizes its convolution for the toggle.

\paragraph{Pre-Volta support.} The default Triton backward for the RWKV-7-style scan emits \texttt{.acq\_rel} atomics that \texttt{ptxas} rejects below \texttt{sm\_70}. An atomic-free two-pass backward reproduces the reference scan's forward and gradients and runs on Pascal (\texttt{sm\_61}) at ${\approx}26\times$ the pure-PyTorch reference throughput (${\approx}2{,}980$ vs.\ ${\approx}113$ tok/s at $d{=}256$/6 layers/ctx 512), with an \texttt{--fp32} flag for hardware without bf16. The headline experiments in this paper were run on a single GTX 1070.

\paragraph{Determinism and floors.} Runs are seed-pinned and resumable; every task ships its floor calculators (value-marginal, no-binding, table floor $1/K$, positional). The rebinding control is validated by a self-test (\texttt{--selftest}) that confirms the derangement logic.

\section{Reproduction Commands}\label{app:commands}

Constrained-state factorial and $K$-sweep (Tables~\ref{tab:k32},~\ref{tab:ksweep}). \texttt{--fp32} is required on pre-bf16 (Pascal) hardware and may be dropped on Ampere and newer; it applies to every command below except \texttt{dg\_c9\_harden.py}, which has no such flag:
\begin{quote}\small\ttfamily
python scripts/mqar\_probe.py --models armed\_gdn gated\_deltanet deltanet \\
\hspace*{1.5em}dg\_rank\_ref dg\_diag\_ref mamba2\_ref mamba2\_ref\_noconv attention \\
\hspace*{1.5em}--d-model 32 --layers 1 --mamba2-d-state 16 --ks 8 16 32 --seeds 0 1 2 --fp32
\end{quote}

Hardened transition ablation with rebinding control (Table~\ref{tab:hardened}):
\begin{quote}\small\ttfamily
python scripts/dg\_c9\_harden.py --ks 16 32 --seeds 0 1 2 3 4 5 6 7 8 9 \\
python scripts/dg\_c9\_harden.py --selftest
\end{quote}

Haystack wall (Table~\ref{tab:wall}):
\begin{quote}\small\ttfamily
python scripts/long\_retrieval\_probe.py --models armed\_gdn dg\_rank\_ref mamba2\_ref attention \\
\hspace*{1.5em}--seq-lens 64 128 256 512 --pairs 4 --seeds 0 1 --heads 2 --mamba2-d-state 16 --fp32
\end{quote}

Curriculum and collision-key discriminator (Tables~\ref{tab:curriculum},~\ref{tab:collision}); add \texttt{--collision} for the discriminator:
\begin{quote}\small\ttfamily
python scripts/long\_retrieval\_probe.py --models armed\_gdn armed\_gdn\_fwd attention \\
\hspace*{1.5em}--seq-lens 64 --pairs 4 --seeds 0 --steps 3000 --batch 256 --layers 2 --heads 2 \\
\hspace*{1.5em}--dense 4 --curriculum --fp32
\end{quote}

$S_5$ guardrail (Table~\ref{tab:s5}):
\begin{quote}\small\ttfamily
python scripts/state\_tracking\_probe.py --models armed\_gdn gated\_deltanet \\
\hspace*{1.5em}--group-n 5 --ls 8 16 --d-model 128 --seeds 0 1 2 3 4 5 6 7 8 9 --fp32
\end{quote}

\section{Floors and Controls, Stated}\label{app:floors}

For each task we report: (i) the \textbf{value-marginal} (probability of the correct value under the answer distribution: $1/26\approx0.038$ for MQAR; $1/54\approx0.019$ for haystack/collision; $1/120\approx0.008$ for $S_5$, exact because generators are uniform over the full group); (ii) the strongest \textbf{no-binding strategy} (MQAR: emit a random table value, ${\approx}0.068$; collision: emit a random table-\emph{position} token, ${\approx}0.26$); (iii) the \textbf{table floor} $1/K$ for the rebinding control. No headline number in this paper is within $10\times$ of its floor except where explicitly marked as ``chance.'' The $S_5$ probe additionally masks all product slots simultaneously and uses single-token answers, making prefix-completion leakage structurally impossible.

\end{document}